\documentclass[letterpaper,10 pt,conference]{ieeeconf}
\IEEEoverridecommandlockouts
\usepackage{ieee_robotics}
\newcommand{\method}{Ag2Manip\xspace}
\newcommand{\methodFullName}{\underline{Ag}ent-\underline{Ag}nostic representations for \underline{Manip}ulation\xspace}
\newcommand{\stageOne}{\textit{exploration}\xspace}
\newcommand{\stageTwo}{\textit{interaction}\xspace}

\usepackage{mwe}
\usepackage{placeins}
\usepackage{progressbar}
\usepackage{ifthen}
\usepackage{xfrac}
\usepackage{dblfloatfix}

\NewDocumentCommand{\mkProgBar}{ O{false} O{red} m }{
  \pgfmathsetmacro\percentage{#3}
  \pgfmathsetmacro\proportion{#3/100}
  \ifthenelse{\equal{#1}{true}} %
    {\underline{\textbf{\percentage\%}}} %
    {\percentage\%} %
  \progressbar[heightr=1.0,width=2.5em,filledcolor=#2!100]{\proportion}%
}

\NewDocumentCommand{\mkCharTB}{ m }{
    \textbf{\texttt{#1}}
}
\newcommand{\lsfrac}[2]{%
  {\LARGE\sfrac{$#1~$}{$~#2$}} 
}
\newcommand{\lsfracbf}[2]{%
  {\LARGE\sfrac{$\mathbf{#1~}$}{$\mathbf{~#2}$}} 
}

\acrodef{ddpm}[DDPM]{Denoising Diffusion Probabilistic Model}
\acrodef{mala}[MALA]{Metropolis-Adjusted Langevin Algorithm}
\acrodef{sdf}[SDF]{Signed Distance Function}
\acrodef{ibs}[IBS]{Intersection Bisector Surface}
\acrodef{fkine}[FK]{Forward Kinematics}
\acrodef{ik}[IK]{Inverse Kinematics}
\acrodef{ppo}[PPO]{Proximal Policy Optimization}
\acrodef{rl}[RL]{Reinforcement Learning}
\acrodef{gcrl}[GCRL]{Goal-Conditioned Reinforcement Learning}
\acrodef{mdp}[MDP]{Markov Decision Process}
\acrodef{pd}[PD]{proportional-derivative}
\acrodef{ood}[OOD]{Out-of-Domain}
\acrodef{dfc}[DFC]{Differentiable Force Closure}
\acrodef{ar}[AR]{Augmented Reality}
\acrodef{llm}[LLM]{Large Language Model}
\acrodef{rwr}[RWR]{Reward-Weighted Regression}

\title{\LARGE \bf
\method: Learning Novel Manipulation Skills\\with Agent-Agnostic Visual and Action Representations\vspace{-12pt}}

\author{Puhao Li$^{1,2,\star}$, Tengyu Liu$^{1,\star}$, Yuyang Li$^{1,2,3}$, Muzhi Han$^{4}$, Haoran Geng$^{1,5}$, Shu Wang$^{4}$,\\
Yixin Zhu$^{3}$, Song-Chun Zhu$^{1,2,3}$, and Siyuan Huang$^{1,\dagger}$\vspace{3pt}\\
\href{https://xiaoyao-li.github.io/research/ag2manip/}{\texttt{xiaoyao-li.github.io/research/ag2manip}}
\thanks{$^\star$ Puhao Li and Tengyu Liu contributed equally to this paper.}
\thanks{$^\dagger$ Corresponding email: {\tt{syhuang@bigai.ai}}.}%
\thanks{$^1$ National Key Laboratory of General Artificial Intelligence, Beijing Institute for General Artificial Intelligence (BIGAI).
$^2$ Department of Automation, Tsinghua University.
$^3$ Institute for Artificial Intelligence, Peking University.
$^4$ University of California, Los Angeles.
$^5$ School of Electronics Engineering and Computer Science, Peking University.%
}
\thanks{This research is in part supported by the National Science and Technology Major Project (2022ZD0114900) and the Beijing Nova Program.}
}%

\begin{document}

\let\oldtwocolumn\twocolumn
\renewcommand\twocolumn[1][]{%
    \oldtwocolumn[{#1}{
        \centering
        \vspace{-12pt}
        \includegraphics[width=\linewidth]{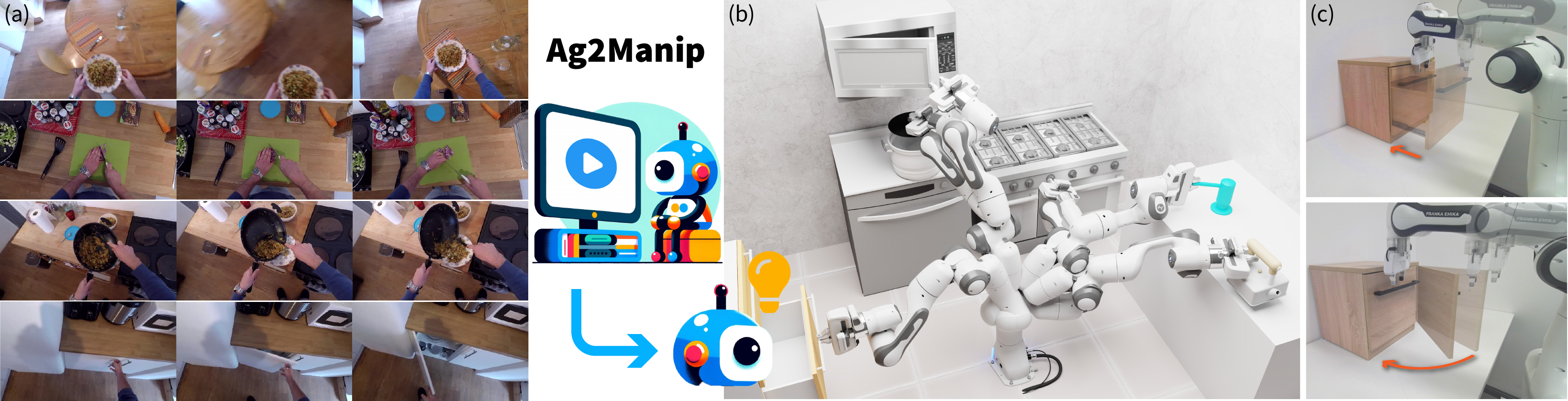}
        \captionof{figure}{\textbf{\method enables various manipulation tasks in scenarios where domain-specific demonstrations are unavailable.} Leveraging agent-agnostic visual and action representations, \method (a) learns from human manipulation videos, removing the reliance on domain-specific examples; (b) autonomously acquires diverse manipulation skills in simulation; and (c) facilitates robust imitation learning of manipulation skills in the real world, demonstrating the practical applicability and generalizability of our approach.}
        \label{fig:teaser}
        \vspace{6pt}
    }]
}

\maketitle
\thispagestyle{empty}
\pagestyle{empty}

\begin{abstract}
Autonomous robotic systems capable of learning novel manipulation tasks are poised to transform industries from manufacturing to service automation. However, modern methods (\eg, VIP and R3M) still face significant hurdles, notably the domain gap among robotic embodiments and the sparsity of successful task executions within specific action spaces, resulting in misaligned and ambiguous task representations. We introduce \method (\methodFullName), a framework aimed at surmounting these challenges through two key innovations: a novel agent-agnostic \textit{visual representation} derived from human manipulation videos, with the specifics of embodiments obscured to enhance generalizability; and an agent-agnostic \textit{action representation} abstracting a robot's kinematics to a universal agent proxy, emphasizing crucial interactions between end-effector and object. \method's empirical validation across simulated benchmarks like FrankaKitchen, ManiSkill, and PartManip shows a 325\% increase in performance, achieved without domain-specific demonstrations. Ablation studies underline the essential contributions of the visual and action representations to this success. Extending our evaluations to the real world, \method significantly improves imitation learning success rates from 50\% to 77.5\%, demonstrating its effectiveness and generalizability across both simulated and physical environments.
\end{abstract}

\section{Introduction}

Robotic systems' capability to autonomously learn and execute novel manipulation skills, without reliance on expert demonstrations, is pivotal as they adapt to changing tasks and environments. Although there have been considerable strides in the domain of learning manipulation skills \cite{brohan2022rt,bahl2022human,zitkovich2023rt,padalkar2023open,wang2023mimicplay,geng2023partmanip,yang2023learning}, the challenge of autonomously acquiring these skills, devoid of expert guidance and task-specific rewards, persists. In addressing this issue, prior research \cite{nair2023r3m,ma2022vip,ma2023eureka} has investigated the use of extensive pre-training to enhance manipulation learning. Notably, recent studies \cite{nair2023r3m,ma2022vip} have focused on developing comprehensive visual representations from human-centric video datasets \cite{damen2020epic,grauman2022ego4d}. These datasets are instrumental in capturing the quintessence of tasks and the temporal dynamics between visual frames, subsequently facilitating the generation of rewards that orient robots towards fulfilling specified objectives. Alternatively, other methodologies \cite{ma2023eureka} incorporate \acp{llm} to directly craft reward functions that aid in the mastery of new manipulation skills. Despite these advancements, existing strategies often falter when confronted with intricate tasks, highlighting three principal challenges in the realm of novel skill acquisition.

First, \textbf{visual representations} derived from human-centric demonstrations \cite{nair2023r3m,ma2022vip} encounter challenges in bridging the gap between the varied appearances and kinematic discrepancies of humans and robots. The appearance discrepancy introduces biases when applied to robots, undermining the models' capacity to accurately decode tasks and their temporal sequences. \textbf{Kinematic differences}, on the other hand, lead to divergent execution strategies; robots might follow trajectories that differ markedly from those in human demonstrations to accomplish tasks like picking up a cup. This variance can cause the model to erroneously classify a robot's optimal path as incorrect due to its reliance on human-centric training data.

Second, the omnipresence of \textbf{human hands} in the training data \textbf{biases} these models towards prioritizing hand appearance, focusing on their position and movement over the actual task objective. For example, in tasks involving cup manipulation, the model may highlight the upward movement of the hands rather than ensuring the cup has been successfully grasped.

Last, the \textbf{demand for precision} in robotic manipulation exacerbates these challenges. Minor trajectory deviations can result in significant performance degradation. While expert-designed rewards provide detailed guidance, those derived from visual or linguistic models are often too broad and high-level, leading to inaccuracies. This issue is particularly acute in tasks requiring precise interaction with the environment, such as opening a door, where exact actions like grasping the handle are essential.

We introduce \textbf{\method}: \methodFullName to address the challenges outlined above. As depicted in \cref{fig:pipeline}, \method features two primary components of generalizable visual and action representations.

To counteract the biases stemming from human-centric training data, we devise an \textbf{agent-agnostic visual representation}. Drawing inspiration from Bahl~\etal~\cite{bahl2022human}, we isolate and obscure both humans and robots within video frames, subsequently inpainting the videos. Training on these agent-obscured frames, in the vein of R3M~\cite{nair2023r3m}, our visual representation transcends the domain gap between humans and robots, fostering robust adaptation to robot-centric tasks. This agent-agnostic visual model prioritizes task processes over human-specific cues, thus providing clearer, more task-focused guidance for manipulation learning.

To mitigate inaccuracies stemming from visual guidance, we propose an \textbf{agent-agnostic action representation}. This framework abstracts robot actions into a universal proxy agent, equipped with a universally applicable action space. This representation divides manipulation learning into two phases: \stageOne and \stageTwo. In \stageOne, the focus is on learning the proxy's trajectory, akin to the end-effector's movements, to enhance environment exploration. Transitioning to \stageTwo when the proxy nears an object's actionable zone shifts the focus to understanding the proxy's exerted forces, simulating end-effector and object interactions. This bifurcation simplifies the learning process, reducing the complexities associated with direct robot and object manipulation. By employing this agent-agnostic action space, our method streamlines task learning, concentrating on pivotal task elements and diminishing the repercussions of sparse guidance. We further complement these representations with a well-structured reward function for each learning stage, fostering interaction and facilitating the translation of learned skills to actual robot arm movements.

\method's effectiveness is showcased through goal-conditioned novel skill learning without expert demonstrations or task-specific rewards, across a variety of simulated tasks in FrankaKitchen~\cite{gupta2019relay}, ManiSkill~\cite{gu2023maniskill2}, and PartManip~\cite{geng2023partmanip}. Our method achieves an impressive \textbf{78.7\%} success rate, significantly outperforming baseline methods with an 18.5\% success rate. By leveraging agent-agnostic visual and action representations, \method significantly advances manipulation learning, equipping robots to adeptly navigate novel tasks in varied environments. Further validation in real-world experiments demonstrates the model's superior skill acquisition capabilities.

In summary, our work introduces three pivotal contributions to the field of learning novel manipulation skills \textbf{without expert input}: (i) an agent-agnostic visual representation that effectively narrows the embodiment gap, enhancing robotic systems' visual data interpretation; (ii) an agent-agnostic action representation that simplifies complex robot actions into more generalizable proxy-agent actions, augmented by a targeted reward function to encourage environmental interaction; and (iii) substantial progress in robot novel skill learning performance, validated across challenging tasks and affirming our approach's practical benefits in boosting robotic adaptability and autonomy.

\section{Related Works}

\subsection{Learning Robotic Manipulation}

The domain of robotic manipulation encompasses both foundational motor skills such as grasping \cite{xu2023unidexgrasp,li2024grasp,li2023gendexgrasp} and manipulation \cite{chen2023visual,chen2023bi,geng2023partmanip,zhao2024tac,geng2022gapartnet}, as well as advanced cognitive capabilities for understanding task specifics, including the location, method, and reasoning behind various tasks \cite{billard2019trends,zhu2020dark,kroemer2021review}. The advent of parallel simulation environments \cite{xiang2020sapien,makoviychuk2021isaac} has facilitated the learning of such skills, though this often necessitates manually tailored reward functions for each task \cite{xu2023unidexgrasp,li2024grasp,chen2023bi}, despite assistance from \acp{llm} and human feedback \cite{ma2023eureka}. A viable alternative involves learning from demonstrations, which bypasses the need for extensive exploration and eases scalability challenges \cite{kroemer2021review}. Robot action trajectories can be captured through teleoperation \cite{qin2023anyteleop,qin2022one}, \ac{ar} systems \cite{duan2023ar2}, and teach pendant programming \cite{brohan2022rt,zitkovich2023rt,padalkar2023open}. However, learning from human videos is a cost-effective yet formidable approach to translating observed interactions into motor controls \cite{qin2022dexmv,qin2022one}, where balancing data collection costs against the quality of demonstrations poses a substantial hurdle in the direct acquisition of new skills from such sources. Inspired by recent advancements \cite{nair2023r3m,ma2022vip}, our study introduces generalizable \textbf{visual and action representations} for the learning of novel manipulation skills across varied tasks, leveraging the wealth of human demonstrations. This strategy seeks to address the challenges inherent in learning directly from videos, presenting a scalable and efficient solution for robotic systems to assimilate new capabilities.

\begin{figure*}[t!]
    \centering
    \includegraphics[width=\linewidth]{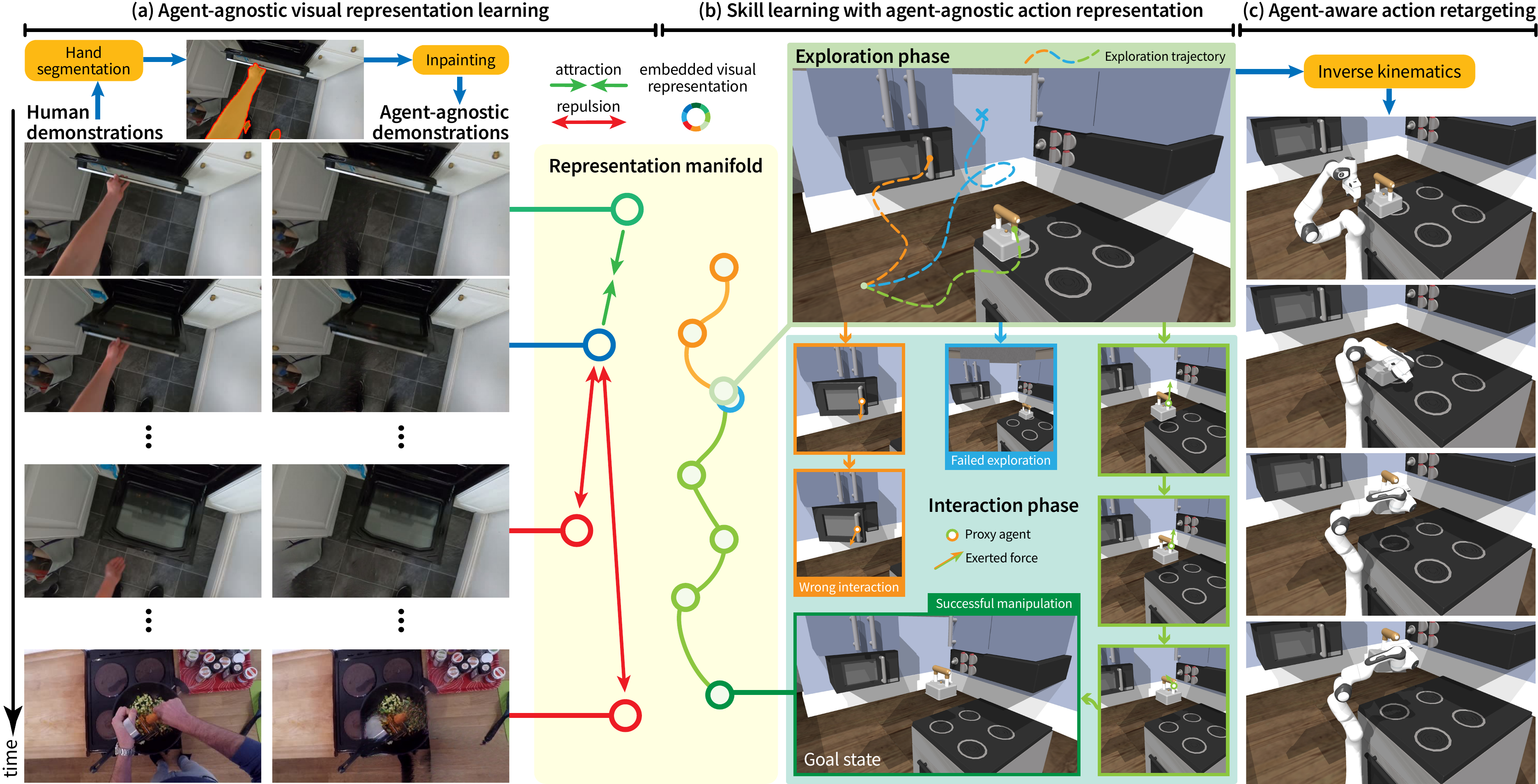}
    \caption{\textbf{Framework of \method.}  Our approach is structured into three primary components: (a) learning an agent-agnostic visual representation, (b) learning abstracted skills via an agent-agnostic action representation, and (c) retargeting the abstracted skills to a robot.}
    \label{fig:pipeline}
\end{figure*}

\subsection{Reward Generation for Skill Learning}

Model-free \ac{rl} for skill learning is notably resource-intensive, primarily due to the necessity for expert-crafted, task- and embodiment-specific rewards. Addressing this issue involves devising an autonomously generated reward function for decision-making pertinent to each task. Foundation models, such as \acp{llm}, have shown potential in directly creating reward functions from task descriptions \cite{fan2022minedojo,kwon2023reward,du2023guiding,ma2023eureka}. However, their effectiveness is somewhat limited without environmental context, often requiring expert feedback to bridge this gap \cite{ma2023eureka}. Additionally, this method's dependency on environmental states, which are usually not readily available in real-world settings, poses a significant challenge. An alternative, perceptual rewards, emerge as a promising avenue for skill learning. By observing human-executed task videos \cite{damen2020epic}, robots can derive an implicit embedding that captures the sequential unfolding of events, serving as a versatile reward mechanism \cite{sermanet2016unsupervised,schmeckpeper2020reinforcement,nair2023r3m}. Advancing this concept, some researchers suggest learning temporal dynamics not from task-specific footage but from a diverse array of tasks, aiming to establish a task-agnostic visual representation with enhanced generalizability \cite{ma2022vip}. Our research builds on these innovations, stripping agent-specific information from the visual reward to further boost its robustness and applicability across various contexts.

\subsection{Agent-Agnostic Representation}

The concept of crafting \textit{agent-agnostic} representations for actions, objects, and tasks serves to abstract them away from the specificities of robotic articulations or sensory setups. This approach significantly boosts adaptability and transferability across different robotic systems and even into human contexts, by separating low-level perceptual and control details in favor of focusing on high-level action abstractions. Such a framework allows for a manipulation task to be conceptualized as the desired alterations in the world state over time, minimizing the agent's direct engagement \cite{chang2024look}. To aptly capture the nuances of agent-object interactions while maintaining agent-agnosticism, the concepts of interaction regions (often correlated with affordances) and trajectories come into play \cite{bahl2022human,wu2022vat,xu2022universal,jiang2022ditto,bharadhwaj2023zero,li2023gendexgrasp,zhang2023flowbot++}. These elements illustrate task execution modalities independent of a robot's specific motor capabilities. For representing interaction zones, a straightforward yet efficacious method involves utilizing contact points to delineate essential contacts between a manipulator (\eg, a finger) and an object \cite{shao2020unigrasp,liu2021synthesizing,jiang2022ditto,zhang2023flowbot++}, catering well to simplistic end-effectors like parallel grippers or suction cups. In scenarios characterized by contact-rich interactions, the adoption of contact maps is indispensable for detailing the extensive contact dynamics or for accurately charting the proximity of each finger to the object surface \cite{brahmbhatt2019contactgrasp,li2023gendexgrasp}.

\section{Method}

This work delves into robotic manipulation learning in scenarios devoid of expert demonstrations. Our objective is to learn robot motions to accomplish a specified goal, given only the robot and an image of the desired end state. To this end, we introduce \textbf{\method}: \methodFullName, whose framework is depicted in \Cref{fig:pipeline}. Our methodology is underpinned by two core innovations: an agent-agnostic visual representation (\cref{sec:method:visual-representation}) that mitigates the domain disparity between humans and robots, and an agent-agnostic action representation (\cref{sec:method:action-representation}) that distills robot actions to those of a universal proxy agent. These foundations enable us to harness \ac{rl} to formulate a manipulation policy within this generalized action space, informed by a novel reward function emerging from our agent-agnostic visual paradigm (\cref{sec:method:reward-shaping}). Subsequently, the trajectory devised for the proxy agent is adapted to the robot through \ac{ik} (\cref{sec:method:robo-act-retarget}), ensuring the practical applicability of the learned maneuvers.

\subsection{Agent-Agnostic Visual Representation}\label{sec:method:visual-representation}

Our work seeks to develop an agent-agnostic visual representation that transcends the domain gap between human and robot manipulations, building on pre-trained visual representations on human demonstrations \cite{nair2023r3m,ma2022vip}. This approach aims to augment the versatility and effectiveness of these representations within robotic contexts, facilitating a more adaptable skill acquisition process.

\paragraph*{Data pre-processing}

We consider a set of human demonstration video data $\mathcal{D} = \{ v^c \coloneqq (o_1^c, o_2^c, ..., o_{n_c}^c) \}_{c=1}^{N}$, where $o_f^c \in \mathbb{R}^{H\times W\times 3}$ is the $f$-th raw frame in the $c$-th video clip $v^c$ that describes how a human completes a manipulation task. Inspired by Bahl \etal~\cite{bahl2022human}, we initiate this process by segmenting the human body from each frame using the ODISE algorithm~\cite{xu2023open}. Following segmentation, we employ a video inpainting model, E$^2$FGVI~\cite{li2022towards}, to fill in the areas previously occupied by the human. This approach not only removes the human from the video but also ensures a smooth temporal coherence between frames, resulting in a manipulation dataset $\mathcal{D}^a$ that is effectively agent-agnostic.

\paragraph*{Time-contrastive pre-training}

Given the agent-agnostic demonstration dataset $\mathcal{D}^a$, we aim to learn an encoder $\mathcal{F}_\phi \colon \mathbb{R}^{H\times W \times 3} \to \mathbb{R}^{K}$ that maps a visual observation into a latent embedding, where $K$ denotes the embedding dimension. Following Nair~\etal~\cite{nair2023r3m}, we minimize the time-contrastive loss~\cite{sermanet2018time} $\mathcal{L}_{\mathrm{tcn}}$ and the regularization penalty $\mathcal{L}_\mathrm{reg}$:
\begin{equation}
    \small
    \mathcal{L} = \lambda_1 \mathbb{E}_{o_i^c,o_j^c,o_k^c,o_l^{\neq c}\sim\mathcal{D}^a} \mathcal{L}_{\mathrm{tcn}} + \lambda_2 \mathbb{E}_{o\sim\mathcal{D}^a} \mathcal{L}_\mathrm{reg},
\end{equation}
where $(o_i^c, o_j^c, o_k^c)\sim v^c$ indicates a set of temporally ordered 3-frame samples, and each sample in a set is drawn from the same video clip $v^c$ to ensure task proximity. $o_l^{\neq c}$ is a negative sample from a disparate video clip. 

The time-contrastive loss is designed to guide the representation so that frames temporally closer to each other are mapped closer in the embedding space, compared to frames that are temporally distant or from disparate video clips:
\begin{equation}
    \small
    \mathcal{L}_\mathrm{tcn} = -\log \frac{e^{\mathcal{S}\left(z_{i}^{c}, z_{j}^{c}\right)}}{e^{\mathcal{S}\left(z_{i}^{c}, z_{j}^{c}\right)}+e^{\mathcal{S}\left(z_{i}^{c}, z_{k}^{c}\right)}+e^{\mathcal{S}\left(z_{i}^{c}, z_{l}^{\neq c}\right)}},
    \label{eq:time-contrastive-loss}
\end{equation}
where $\mathcal{S}(\cdot, \cdot)$ represents the similarity metric between two embeddings, $z_i^c = \mathcal{F}_\phi(o_i^c)$ denotes the embedding of $o_i^c$ extracted from the encoder $\mathcal{F}_\phi$. The regularization loss encourages a more compact embedding space:
\begin{equation}
    \small
    \mathcal{L}_\mathrm{reg} = \Vert \mathcal{F}_\phi(o) \Vert_1 + \Vert \mathcal{F}_\phi(o) \Vert_2 .
\end{equation}

\subsection{Agent-Agnostic Action Representation}\label{sec:method:action-representation}

Our approach encapsulates robotic manipulation learning via an agent-agnostic action representation. This involves abstracting a robot's movements and interactions into those of a universal, free-floating proxy agent, encompassing both motion and exerted forces. The learning process is bifurcated into two stages: \stageOne, concentrating on the proxy's positional adjustments, and \stageTwo, focusing on the forces applied by the proxy on the environment. An \ac{rl} policy is developed to minimize the embedding distance in the agent-agnostic visual representation space between the current state and a goal state depicted by an image.

\paragraph*{The \stageOne phase}

The robot is abstracted as a universal proxy agent, represented by an agent-agnostic sphere to mimic the end-effector's actions, translating the robot's actions into a sequence of positions for this sphere. Control over the proxy is established through a \ac{pd} controller \cite{tan2011stable}, with the proxy embodying a collision volume of radius $r_\mathrm{e}$ to denote its physical presence. This phase concludes upon the proxy's arrival at a precalculated interactable region within the environment, marking the commencement of the \stageTwo phase. For the scope of this work which focuses on robots equipped with two-finger grippers, interactable regions are identified as zones where parallel grips are deemed feasible. Utilizing point cloud scans of the environment, these regions are determined based on proximity to potential gripping points identified by GraspNet \cite{fang2020graspnet}. Although parallel grip detection is utilized for its efficiency in our setup, general-purpose methods like GenDexGrasp \cite{li2023gendexgrasp} could also delineate interactable regions suitable for a range of dexterous manipulations.

\paragraph*{The \stageTwo phase}

With the proxy's entry into an interactable region, indicating a viable grasp and subsequent object attachment, the focus shifts to the \stageTwo phase. This stage is dedicated to the manipulation of the object, abstracting the robot's actions into the forces the proxy exerts upon the environment.

\begin{table*}[b!]
    \centering
    \small
    \setlength{\tabcolsep}{3pt}
    \caption{\textbf{Comparisons and ablation studies}. Each task was evaluated over $3~\text{seeds} \times 3~\text{cameras} = \mathbf{9}~\textbf{runs}$, with the numbers $\mathbf{0-9}$ indicating the count of successful attempts. The characters \textbf{\texttt{a - x}} denote specific tasks. Tasks from FrankaKitchen~\cite{gupta2019relay} include: \mkCharTB{a:} open hinge-cabinet, \mkCharTB{b:} open microwave, \mkCharTB{c:} open slide-cabinet, \mkCharTB{d:} close hinge-cabinet, \mkCharTB{e:} close microwave, \mkCharTB{f:} close slide-cabinet, \mkCharTB{g:} move kettle, \mkCharTB{h:} pick up kettle, \mkCharTB{i:} turn on switch, and \mkCharTB{j:} turn off switch. Tasks from ManiSkill2~\cite{gu2023maniskill2} include: \mkCharTB{k:} open door, \mkCharTB{l:} close door, \mkCharTB{m:} pick up cube, \mkCharTB{n:} stack cube, \mkCharTB{o:} pick up clutterycb, \mkCharTB{p:} insert peg, \mkCharTB{q:} turn left faucet, and \mkCharTB{r:} turn right faucet. Tasks from PartManip~\cite{geng2023partmanip} include: \mkCharTB{s:} turn down dishwasher, \mkCharTB{t:} pull drawer, \mkCharTB{u:} turn up dishwasher, \mkCharTB{v:} push drawer, \mkCharTB{w:} press button, and \mkCharTB{x:} lift lid.}
    \label{tab:main_com}
    \resizebox{\linewidth}{!}{%
        \begin{tabular}{ccccccccccccccccccccccccccccc}%
            \toprule
            \multirow{2}{*}{\textbf{Method}} & \multicolumn{11}{c}{\textbf{FrankaKitchen}} & \multicolumn{9}{c}{\textbf{ManiSkill}} & \multicolumn{7}{c}{\textbf{PartManip}} & \multirow{2}{*}{\textbf{Overall}} \\
            \cmidrule(lr){2-12}\cmidrule(lr){13-21}\cmidrule(lr){22-28} & \mkCharTB{a} & \mkCharTB{b} & \mkCharTB{c} & \mkCharTB{d} & \mkCharTB{e} & \mkCharTB{f} & \mkCharTB{g} & \mkCharTB{h} & \mkCharTB{i} & \mkCharTB{j} & \textbf{Avg.} & \mkCharTB{k} & \mkCharTB{l} & \mkCharTB{m} & \mkCharTB{n} & \mkCharTB{o} & \mkCharTB{p} & \mkCharTB{q} & \mkCharTB{r} & \textbf{Avg.} & \mkCharTB{s} & \mkCharTB{t} & \mkCharTB{u} & \mkCharTB{v} & \mkCharTB{w} & \mkCharTB{x} & \textbf{Avg.} &  \\
            \midrule
            R3M~\cite{nair2023r3m} & 0 & 0 & 0 & 3 & 2 & 0 & 1 & 0 & 0 & 0 & 6.7\% & 0 & 6 & 0 & 0 & 0 & 0 & 0 & 0 & 8.3\% & 0 & 0 & 3 & 9 & 0 & 0 & 22.2\% & \mkProgBar[false][red]{11.1} \\
            VIP~\cite{ma2022vip} & 0 & 0 & 0 & 2 & 6 & 0 & 3 & 0 & 0 & 0 & 12.2\% & 0 & 6 & 0 & 0 & 0 & 0 & 0 & 0 & 8.3\% & 0 & 0 & 0 & 9 & 0 & 0 & 16.7\% & \mkProgBar[false][red]{12.0} \\
            Eureka~\cite{ma2023eureka} & 0 & 0 & 0 & 7 & 3 & 2 & 3 & 0 & 0 & 0 & 16.7\% & 0 & 9 & 0 & 0 & 0 & 0 & 0 & 1 & 13.9\% & 0 & 0 & 3 & 6 & 0 & 0 & 20.0\% & \mkProgBar[false][red]{18.5} \\
            \midrule
            Ours w/o Act.Repr. & 4 & 1 & 8 & 9 & 9 & 9 & 9 & 1 & 7 & 2 & 65.6\% & 0 & 9 & 0 & 0 & 0 & 0 & 1 & 8 & 25.0\% & 0 & 0 & 8 & 9 & 0 & 0 & 31.5\% & \mkProgBar[false][orange]{43.5} \\
            Ours w/o Rew.Shp. & 8 & 7 & 7 & 9 & 9 & 9 & 7 & 9 & 1 & 0 & 73.3\% & 9 & 9 & 8 & 0 & 3 & 1 & 4 & 5 & 54.2\% & 9 & 6 & 8 & 9 & 0 & 9 & 75.9\% & \mkProgBar[false][orange]{67.6} \\
            \textbf{Ours} & 7 & 8 & 8 & 8 & 8 & 9 & 8 & 6 & 9 & 9 & \underline{\textbf{88.9\%}} & 7 & 9 & 6 & 0 & 7 & 2 & 8 & 8 & \underline{\textbf{65.3\%}} & 9 & 7 & 9 & 9 & 0 & 9 & \underline{\textbf{79.6\%}} & \mkProgBar[true][OliveGreen]{78.7} \\
            \midrule
            Ours (Proxy) & 8 & 9 & 9 & 8 & 9 & 9 & 9 & 9 & 9 & 9 & 97.8\% & 7 & 9 & 5 & 5 & 7 & 3 & 8 & 9 & 73.6\% & 9 & 9 & 9 & 9 & 0 & 8 & 81.5\% & \mkProgBar[false][OliveGreen]{85.7} \\
            \bottomrule
        \end{tabular}%
    }%
\end{table*}

\subsection{Reinforcement Learning and Reward Shaping}\label{sec:method:reward-shaping}

Given a goal image $g\in\mathbb{R}^{H\times W\times3}$, our task is to accomplish the task it represents. We use a model-free and \ac{gcrl} framework to learn the agent-agnostic action policy $\pi = \{\pi_\mathrm{exp}\text, \pi_\mathrm{int}\}$, with $\pi_\mathrm{exp}$ and $\pi_\mathrm{int}$ denoting the proxy agent's policies for the \stageOne and \stageTwo phases, respectively. The policy $\pi$ takes the robot states $r_t$ and the environment's states $s_t$ at frame $t$ as its observation and produces the action $a_t = (a_p^t, a_f^t)$, where $a_p^t \in \mathbb{R}^3$ indicates the proxy's desired position in \stageOne and $a_f^t \in \mathbb{R}^3$ indicates the proxy's intended force in \stageTwo. A \ac{pd} controller then guides the proxy to achieve the target action.

To reach the goal depicted by $g$, we focus on maximizing the similarity $\mathcal{S}(z_t,z_g)$ between the embeddings for current and goal images $o_t$ and $g$. Recognizing that directly employing $\mathcal{S}$ as a reward function could inappropriately penalize trajectories close but not identical to optimal, we introduce an importance-weighted reward function to promote explorations leading to states that improve upon the initial state:
\begin{equation}
    \small
    \resizebox{.91\linewidth}{!}{%
        $\displaystyle
        \mathcal{R}(o_t, g; \phi) = \exp\left( \left( 1 + \alpha \cdot \mathbf{1}_{\mathcal{S}(z_t, z_g) - \beta > 0} \right) \frac{\mathcal{S}(z_t, z_g) - \beta}{\beta} \right) - 1,
    $}%
\end{equation}
where $\beta = \mathcal{S}(z_0, z_g)$ denotes the similarity between the embeddings of the start and goal images, and $\alpha>0$ is an tunable hyperparameter. This reward function, with its indicator function, prioritizes states closer to the goal relative to the starting point and lessens the penalty for deviations, thus promoting exploration beneficial in the policy's early phase of learning with random policy behaviors.

For policy optimization, we utilize \ac{ppo}~\cite{schulman2017proximal}, chosen for its training stability and efficiency in convergence. Through \ac{ppo}, we aim to maximize the expected cumulative reward $\mathbb{E}\left[ \sum_{t=0}^{T-1} \gamma^t \mathcal{R}(o_t, g; \phi) \right]$, thereby effectively guiding the policy $\pi$ towards the goal.

\subsection{Robot-Specific Action Retargeting}\label{sec:method:robo-act-retarget}

To facilitate the transition of the proxy's trajectory, as determined by $\pi$, into actionable movements for real robots, we employ a retargeting policy that translates proxy actions into robot-specific actions. During the \stageOne phase, the positions of the proxy agent are directly mapped to the robot's end-effector positions, thereby converting the proxy's navigational path into corresponding end-effector motions. As the process shifts from \stageOne to \stageTwo, the end-effector's 6D pose is adjusted to align with the nearest viable grasp pose as identified by GraspNet, an approach that is feasible because this transition is predicated on the proximity of an achievable grasp. In the \stageTwo phase, the movement of the object dictates the end-effector's 6D trajectory, ensuring the robot's actions remain in harmony with the object's dynamics. The trajectory for the robot arm is calculated using \ac{ik}, aligning the practical task execution with the proxy's intended actions.

\subsection{Implementation Details}

In \cref{sec:method:visual-representation}, we choose Epic-Kitchen~\cite{damen2020epic} as the human demonstration dataset. Echoing the choices of R3M~\cite{nair2023r3m} and VIP~\cite{ma2022vip}, we use a standard ResNet50~\cite{he2016deep} as the architecture of the visual encoder $\mathcal{F}_\phi$. We use the negative L2 distance to measure similarity $\mathcal{S}(\cdot, \cdot)$. The weights for our learning objective are set to $\lambda_1 = \lambda_2 = 1.0$. The optimization of the visual encoder is carried out using an Adam optimizer with a learning rate of $10^{-4}$, over a duration of 24 hours on a single NVIDIA A100 GPU. In \cref{sec:method:action-representation}, the collision and interactive region radii are defined as 2 centimeters ($r_\mathrm{e}$) and 10 centimeters ($r_\mathrm{int}$). For the reward shaping in \cref{sec:method:reward-shaping}, $\alpha = 3.0$ is empirically determined as the hyperparameter of the reward function across all tasks.

\section{Simulations and Experiments}

Our comprehensive evaluation of the proposed \method showcases significant enhancements in terms of task success rates, achieving a leap from a baseline success rate of 18.5\% to an impressive 78.7\% across tasks sourced from three different environments. Furthermore, our visual representation contributes to a marked increase in the success rate of imitation learning, which increases from 50\% to 77.5\%. These advancements highlight the \method's effectiveness and its considerable promise for real-world applications.

\subsection{Simulation Setup}

\paragraph*{Environments}

To assess the broad applicability of the proposed \method across various manipulation tasks, we select 24 distinct tasks from three varied simulation environments. FrankaKitchen~\cite{gupta2019relay}, ManiSkill~\cite{gu2023maniskill2}, and PartManip~\cite{geng2023partmanip}. These tasks span a wide range of actions, including opening, pulling, and moving, and involve interactions with a variety of objects like cabinets, microwaves, and kettles, executed using a 9-DOF Franka Emika robotic arm and gripper. This setup typifies a standard in robotic manipulation.

Experiments are conducted within the NVIDIA IsaacGym, leveraging its GPU acceleration for efficient \ac{rl}-based learning. The robot initiates each task from a standardized default position, with task objectives defined by goal states represented by images rendered from one of three predetermined camera perspectives (front, left, right). Success in a task is determined by the object or component reaching its goal state within a predefined error margin. To ensure a thorough evaluation, each of the 24 tasks undergoes testing in 9 varied setups combining different camera angles and initialization seeds (3 cameras × 3 seeds), providing a comprehensive overview of performance across multiple conditions.

\begin{figure*}[t!]
    \centering
    \includegraphics[width=\linewidth]{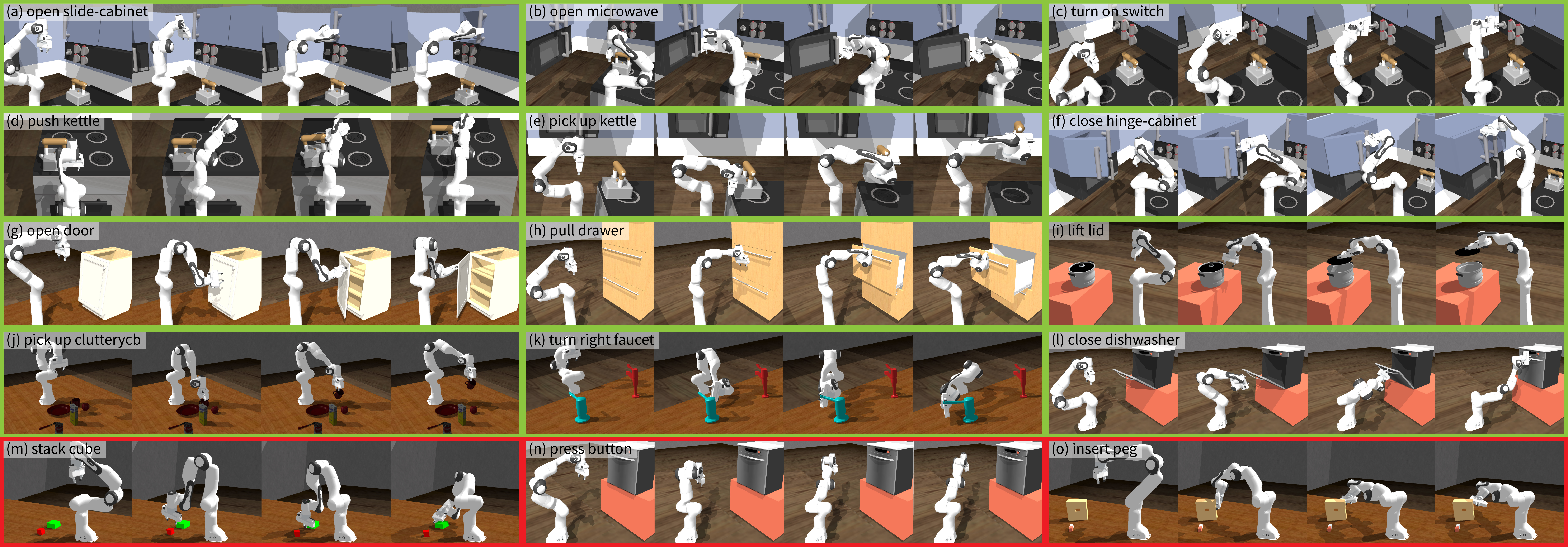}
    \caption{\textbf{Qualitative results in simulation.} The top four rows are successful executions, whereas the bottom row shows failures.}
    \label{fig:setup:qual-res}
\end{figure*}

\paragraph*{Baselines}

Our approach is compared against two baselines, R3M~\cite{nair2023r3m} and VIP~\cite{ma2022vip}, which utilize agent-aware visual representations and time-contrastive learning objectives for learning manipulation skills. Eureka, a novel method distinguished for its ability to autonomously generate reward functions via \acp{llm}, also stands as a significant benchmark and highlights its strengths in skill learning.

For equitable comparison, all methods, barring Eureka, are built upon a ResNet50 architecture and trained using the Epic-Kitchen dataset. To eliminate the influence of task-specific expert insights, Eureka's human feedback feature was deactivated, ensuring that the evaluation focuses solely on each method's intrinsic learning capabilities.

\paragraph*{Ablations}

Our ablation study delineates the impact of distinct components by excluding them from our method. \textit{Ours w/o Act.Repr.} investigates learning directly within the robot's native action space while retaining the agent-agnostic visual representation. Conversely, \textit{Ours w/o Rew.Shp.} employs a straightforward similarity metric instead of our tailored reward function. The removal of solely the visual representation was not considered, given the impracticality of computing agent-aware visuals without corresponding actions. Similarly, excluding both representations would essentially replicate the R3M baseline. Additionally, \textit{Ours (Proxy)} examines the efficacy of the proxy agent's performance devoid of action retargeting to a robot, thereby assessing the impact of retargeting on performance.

\subsection{Results: Simulation}

The summarized results in \cref{tab:main_com} detail the average task success rates within each of the three environments and cumulatively. \method emerges as a standout, securing an overall task success rate of \textbf{78.7\%}, markedly surpassing the baseline methods which recorded success rates of 11.1\%, 12.0\%, and 18.5\%. Further dissecting the success rates per task reveals the distinct competencies of each method. Notably, baseline approaches underperform in tasks demanding precise robot-object interactions, such as door opening or kettle lifting, which require initial attachment actions that often elude the baselines. Eureka exhibits similar shortcomings, which we ascribe to the absence of expert-in-the-loop feedback, consequently affecting its ability to generate refined reward signals. In contrast, \method adeptly acquires these challenging skills, benefitting from its foundational agent-agnostic visual and action representations. 

Nonetheless, \method does encounter consistent challenges with specific tasks: cube stacking, peg insertion, and button pressing. These difficulties arise from a range of issues including collision occurrences with the robot arm in cube stacking, complex object interactions beyond the training set's scope for peg insertion, and the lack of substantial visual cues for button pressing due to minor appearance changes. Potential resolutions could entail integrating more sophisticated planning methods, broadening the scope of human demonstration videos for training the visual representation, and incorporating more guiding elements like the anticipated trajectory of the end-effector to refine task performance.

Additionally, \cref{fig:setup:qual-res} illustrates some of the manipulation trajectories learned by \method, demonstrating its efficacy in handling both rigid and articulated objects across \cref{fig:setup:qual-res} (a-l), and delineating instances of failure in \cref{fig:setup:qual-res} (m-o).

\subsection{Results: Ablation}

Substituting our meticulously crafted reward function with a basic similarity metric (\textit{Ours w/o Rew.Shp.}) led to an 11.1\% reduction in overall task success rates. This significant decline accentuates the pivotal role our reward shaping plays in facilitating the completion of intricate tasks, particularly those necessitating precise movements like turning and lifting. The omission of the agent-agnostic action representation (\textit{Ours w/o Act.Repr.}) had an even more marked effect, with a 35.2\% drop in success, underscoring its critical contribution to \method's performance in tasks that demand accurate control, such as pulling and opening. Notably, even with this reduction, this configuration still outperforms the R3M baseline by 32.4\%, highlighting the value added by our agent-agnostic visual representation.

Examining the performance of our agent-agnostic proxy agent before retargeting its actions to a robot (\textit{Ours (Proxy)}) revealed that the retargeting step accounts for a 7.0\% decrease in success rates. This finding points to the retargeting phase as a promising avenue for further enhancing \method's effectiveness.

\subsection{Visual Representation: Task Progress Consistency}

To verify the consistency of our visual representation in mirroring the progression within a manipulation task, we employed the Spearman Rank Correlation~\cite{spearman1987proof} to analyze expert trajectories. This approach compares the temporal sequence of video frames with their respective similarities to the task's goal state, aiming to ascertain whether initial frames generally exhibit lesser similarity to the goal than subsequent frames, indicative of coherent task advancement.

The proposed \method is benchmarked against several established baselines, such as a ResNet50~\cite{he2016deep} model pre-trained on ImageNet for general image classification, CLIP~\cite{radford2021learning,shah2021rrl}, R3M~\cite{nair2023r3m}, and VIP~\cite{ma2022vip}. These models span a range of applications, from basic image recognition to robotic control tasks, offering a broad spectrum for comparative analysis. The evaluation encompassed 72 expert trajectories---three per task---for the 24 tasks delineated in prior experiments.

According to the results tabulated in \cref{tab:value_rank}, our agent-agnostic visual representation demonstrates a higher consistency with the logical task progression over time, surpassing the baseline models. This implies that our approach provides more accurate and dependable cues for task learning, thereby improving the robot's comprehension and execution of tasks through visual guidance.

\begin{table}[ht!]
    \centering
    \footnotesize
    \setlength{\tabcolsep}{3pt}
    \caption{\textbf{Task progress consistency of visual representation.}} \label{tab:value_rank}
    \resizebox{\linewidth}{!}{%
        \begin{tabular}{ccccc}
            \toprule
            \textbf{Method} & \textbf{FrankaKitchen} & \textbf{ManiSkill} & \textbf{PartManip} & \textbf{Overall} \\
            \midrule
            ResNet50~\cite{he2016deep} & $0.535^{\pm .169}$ & $0.407^{\pm .182}$ & $0.202^{\pm .197}$ & $0.418^{\pm .199}$ \\
            CLIP~\cite{radford2021learning} & $0.627^{\pm .086}$ & $0.381^{\pm .139}$ & $0.347^{\pm .151}$ & $0.490^{\pm .134}$  \\
            R3M~\cite{nair2023r3m} & $0.498^{\pm .190}$ & $0.393^{\pm .191}$ & $0.525^{\pm .123}$ & $0.474^{\pm .177}$  \\
            VIP~\cite{ma2022vip} & $0.496^{\pm .246}$ & $0.251^{\pm .178}$ & $0.386^{\pm .121}$ & $0.401^{\pm .208}$  \\
            \midrule
            \textbf{\method{}} & $\mathbf{0.828^{\pm .082}}$ & $\mathbf{0.696^{\pm .182}}$ & $\mathbf{0.618^{\pm .227}}$ & $\mathbf{0.740^{\pm .153}}$ \\
            \bottomrule
        \end{tabular}
    }%
\end{table}

\begin{figure}[ht!]
    \centering
    \includegraphics[width=\linewidth]{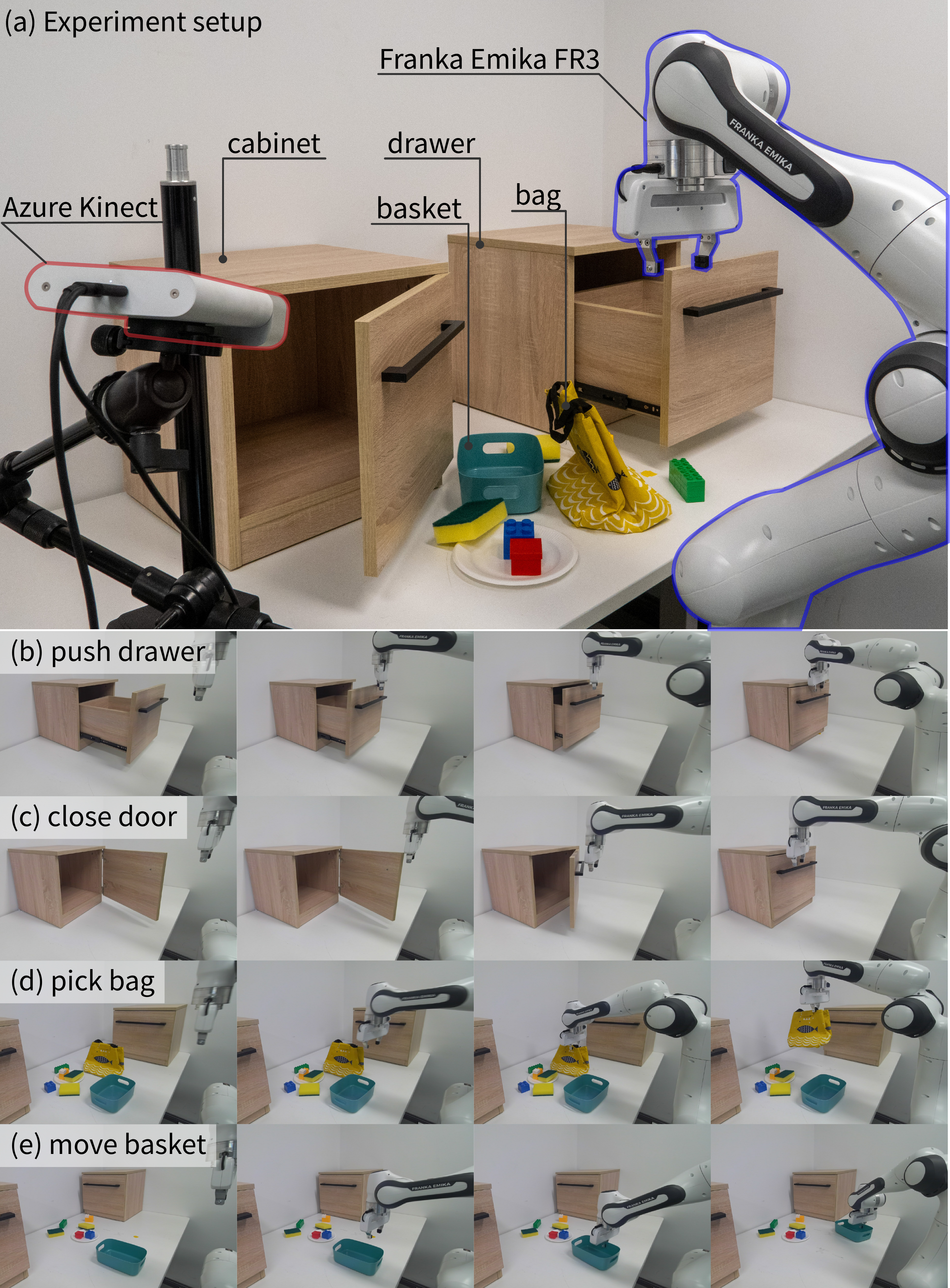}
    \caption{\textbf{Experimental setup.}}
    \label{fig:setup_real}
\end{figure}

\subsection{Visual Representation: Experiments on Imitation}

This experiment aims to evaluate the effectiveness of our visual representation in real-world few-shot imitation learning scenarios. Utilizing a Franka Emika FR3 robot and a Kinect Azure camera, as depicted in \cref{fig:setup_real}, we explore four manipulation tasks: PushDrawer, CloseDoor, PickBag, and MoveBasket. For each task, we gather 20 demonstrations to facilitate the imitation learning process.

We implement advantage-weighted regression~\cite{peng2019advantage} for this experiment, a strategy that accentuates transitions contributing significantly to task progression. This approach assigns weights by assessing the similarity between consecutive observations and the task's goal state, thereby incentivizing actions that evidently advance toward task completion.

The specifics of our experimental setup and the results are shown in \cref{tab:real_imitation,fig:setup_real}. Our findings indicate that the agent-agnostic visual representation notably outperforms the baselines, including ResNet50 and CLIP, which do not undergo task-specific pre-training, as well as R3M and VIP, which exhibit commendable performance barring certain exceptions. Our approach demonstrates a superior capability in narrowing the domain gap that often exists between training datasets and real-world observations, capturing the critical action trajectories necessary for successful task execution within a few-shot learning framework.

\begin{table}[ht!]
    \centering
    \footnotesize
    \setlength{\tabcolsep}{3pt}
    \caption{\textbf{Experimental results.}}\label{tab:real_imitation}
        \begin{tabular}{ccccc}
            \toprule
            \textbf{Method} & PushDrawer & CloseCabinet & PickBag & MoveBasket \\
            \midrule
            ResNet50~\cite{he2016deep}      & \lsfrac{1}{10} & \lsfrac{5}{10} & \lsfrac{1}{10} & \lsfrac{1}{10} \\
            CLIP~\cite{radford2021learning} & \lsfrac{2}{10} & \lsfrac{3}{10} & \lsfrac{0}{10} & \lsfrac{0}{10} \\
            R3M~\cite{nair2023r3m}          & \lsfrac{4}{10} & \lsfrac{5}{10} & \lsfrac{4}{10} & \lsfrac{3}{10} \\
            VIP~\cite{ma2022vip}            & \lsfrac{6}{10} & \lsfrac{6}{10} & \lsfrac{2}{10} & \lsfrac{6}{10} \\
            \midrule
            \textbf{\method{}}              & \lsfracbf{7}{10} & \lsfracbf{8}{10} & \lsfracbf{8}{10} & \lsfracbf{8}{10} \\
            \bottomrule
        \end{tabular}
\end{table}

\section{Conclusion}

In this work, we introduced \method, a novel framework enabling robots to acquire a wide array of manipulation skills without the necessity of expert demonstrations. Our method is grounded in the development of novel agent-agnostic visual and action representations, designed to bridge the domain disparities between various robot embodiments and address the intricate precision requirements inherent in robotic manipulations. Evaluated through extensive simulations and real-world experiments, \method has proven to significantly improve the process of learning robotic manipulation skills, underscoring its effectiveness in facilitating autonomous skill acquisition in robots. This achievement represents a significant leap towards the realization of versatile embodied agents equipped to navigate and adapt to new challenges seamlessly.

{
\bibliographystyle{ieeetr}
\footnotesize
\setstretch{0.92}
\balance
\bibliography{reference_header_shorter,reference}

\begin{thebibliography}{10}

\bibitem{brohan2022rt}
A.~Brohan, N.~Brown, J.~Carbajal, Y.~Chebotar, J.~Dabis, C.~Finn, K.~Gopalakrishnan, K.~Hausman, A.~Herzog, J.~Hsu, {\em et~al.}, ``Rt-1: Robotics transformer for real-world control at scale,'' {\em arXiv preprint arXiv:2212.06817}, 2022.

\bibitem{bahl2022human}
S.~Bahl, A.~Gupta, and D.~Pathak, ``Human-to-robot imitation in the wild,'' in {\em RSS}, 2022.

\bibitem{zitkovich2023rt}
B.~Zitkovich, T.~Yu, S.~Xu, P.~Xu, T.~Xiao, F.~Xia, J.~Wu, P.~Wohlhart, S.~Welker, A.~Wahid, {\em et~al.}, ``Rt-2: Vision-language-action models transfer web knowledge to robotic control,'' in {\em CoRL}, 2023.

\bibitem{padalkar2023open}
A.~Padalkar, A.~Pooley, A.~Jain, A.~Bewley, A.~Herzog, A.~Irpan, A.~Khazatsky, A.~Rai, A.~Singh, A.~Brohan, {\em et~al.}, ``Open x-embodiment: Robotic learning datasets and rt-x models,'' {\em arXiv preprint arXiv:2310.08864}, 2023.

\bibitem{wang2023mimicplay}
C.~Wang, L.~Fan, J.~Sun, R.~Zhang, L.~Fei-Fei, D.~Xu, Y.~Zhu, and A.~Anandkumar, ``Mimicplay: Long-horizon imitation learning by watching human play,'' {\em arXiv preprint arXiv:2302.12422}, 2023.

\bibitem{geng2023partmanip}
H.~Geng, Z.~Li, Y.~Geng, J.~Chen, H.~Dong, and H.~Wang, ``Partmanip: Learning cross-category generalizable part manipulation policy from point cloud observations,'' in {\em CVPR}, 2023.

\bibitem{yang2023learning}
M.~Yang, Y.~Du, K.~Ghasemipour, J.~Tompson, D.~Schuurmans, and P.~Abbeel, ``Learning interactive real-world simulators,'' {\em arXiv preprint arXiv:2310.06114}, 2023.

\bibitem{nair2023r3m}
S.~Nair, A.~Rajeswaran, V.~Kumar, C.~Finn, and A.~Gupta, ``R3m: A universal visual representation for robot manipulation,'' in {\em CoRL}, 2023.

\bibitem{ma2022vip}
Y.~J. Ma, S.~Sodhani, D.~Jayaraman, O.~Bastani, V.~Kumar, and A.~Zhang, ``Vip: Towards universal visual reward and representation via value-implicit pre-training,'' in {\em ICLR}, 2023.

\bibitem{ma2023eureka}
Y.~J. Ma, W.~Liang, G.~Wang, D.-A. Huang, O.~Bastani, D.~Jayaraman, Y.~Zhu, L.~Fan, and A.~Anandkumar, ``Eureka: Human-level reward design via coding large language models,'' {\em arXiv preprint arXiv:2310.12931}, 2023.

\bibitem{damen2020epic}
D.~Damen, H.~Doughty, G.~M. Farinella, S.~Fidler, A.~Furnari, E.~Kazakos, D.~Moltisanti, J.~Munro, T.~Perrett, W.~Price, {\em et~al.}, ``The epic-kitchens dataset: Collection, challenges and baselines,'' {\em TPAMI}, vol.~43, no.~11, pp.~4125--4141, 2020.

\bibitem{grauman2022ego4d}
K.~Grauman, A.~Westbury, E.~Byrne, Z.~Chavis, A.~Furnari, R.~Girdhar, J.~Hamburger, H.~Jiang, M.~Liu, X.~Liu, {\em et~al.}, ``Ego4d: Around the world in 3,000 hours of egocentric video,'' in {\em CVPR}, 2022.

\bibitem{gupta2019relay}
A.~Gupta, V.~Kumar, C.~Lynch, S.~Levine, and K.~Hausman, ``Relay policy learning: Solving long-horizon tasks via imitation and reinforcement learning,'' in {\em CoRL}, 2019.

\bibitem{gu2023maniskill2}
J.~Gu, F.~Xiang, X.~Li, Z.~Ling, X.~Liu, T.~Mu, Y.~Tang, S.~Tao, X.~Wei, Y.~Yao, {\em et~al.}, ``Maniskill2: A unified benchmark for generalizable manipulation skills,'' in {\em ICLR}, 2023.

\bibitem{xu2023unidexgrasp}
Y.~Xu, W.~Wan, J.~Zhang, H.~Liu, Z.~Shan, H.~Shen, R.~Wang, H.~Geng, Y.~Weng, J.~Chen, {\em et~al.}, ``Unidexgrasp: Universal robotic dexterous grasping via learning diverse proposal generation and goal-conditioned policy,'' in {\em CVPR}, 2023.

\bibitem{li2024grasp}
Y.~Li, B.~Liu, Y.~Geng, P.~Li, Y.~Yang, Y.~Zhu, T.~Liu, and S.~Huang, ``Grasp multiple objects with one hand,'' {\em RA-L}, 2024.

\bibitem{li2023gendexgrasp}
P.~Li, T.~Liu, Y.~Li, Y.~Geng, Y.~Zhu, Y.~Yang, and S.~Huang, ``Gendexgrasp: Generalizable dexterous grasping,'' in {\em ICRA}, 2023.

\bibitem{chen2023visual}
T.~Chen, M.~Tippur, S.~Wu, V.~Kumar, E.~Adelson, and P.~Agrawal, ``Visual dexterity: In-hand reorientation of novel and complex object shapes,'' {\em Science Robotics}, vol.~8, no.~84, p.~eadc9244, 2023.

\bibitem{chen2023bi}
Y.~Chen, Y.~Geng, F.~Zhong, J.~Ji, J.~Jiang, Z.~Lu, H.~Dong, and Y.~Yang, ``Bi-dexhands: Towards human-level bimanual dexterous manipulation,'' {\em TPAMI}, 2023.

\bibitem{zhao2024tac}
Z.~Zhao, Y.~Li, W.~Li, Z.~Qi, L.~Ruan, Y.~Zhu, and K.~Althoefer, ``Tac-man: Tactile-informed prior-free manipulation of articulated objects,'' {\em arXiv preprint arXiv:2403.01694}, 2024.

\bibitem{geng2022gapartnet}
H.~Geng, H.~Xu, C.~Zhao, C.~Xu, L.~Yi, S.~Huang, and H.~Wang, ``Gapartnet: Cross-category domain-generalizable object perception and manipulation via generalizable and actionable parts,'' {\em arXiv preprint arXiv:2211.05272}, 2022.

\bibitem{billard2019trends}
A.~Billard and D.~Kragic, ``Trends and challenges in robot manipulation,'' {\em Science}, vol.~364, no.~6446, p.~eaat8414, 2019.

\bibitem{zhu2020dark}
Y.~Zhu, T.~Gao, L.~Fan, S.~Huang, M.~Edmonds, H.~Liu, F.~Gao, C.~Zhang, S.~Qi, Y.~N. Wu, {\em et~al.}, ``Dark, beyond deep: A paradigm shift to cognitive ai with humanlike common sense,'' {\em Engineering}, vol.~6, no.~3, pp.~310--345, 2020.

\bibitem{kroemer2021review}
O.~Kroemer, S.~Niekum, and G.~Konidaris, ``A review of robot learning for manipulation: Challenges, representations, and algorithms,'' {\em JMLR}, vol.~22, no.~1, pp.~1395--1476, 2021.

\bibitem{xiang2020sapien}
F.~Xiang, Y.~Qin, K.~Mo, Y.~Xia, H.~Zhu, F.~Liu, M.~Liu, H.~Jiang, Y.~Yuan, H.~Wang, {\em et~al.}, ``Sapien: A simulated part-based interactive environment,'' in {\em CVPR}, 2020.

\bibitem{makoviychuk2021isaac}
V.~Makoviychuk, L.~Wawrzyniak, Y.~Guo, M.~Lu, K.~Storey, M.~Macklin, D.~Hoeller, N.~Rudin, A.~Allshire, A.~Handa, {\em et~al.}, ``Isaac gym: High performance gpu-based physics simulation for robot learning,'' {\em arXiv preprint arXiv:2108.10470}, 2021.

\bibitem{qin2023anyteleop}
Y.~Qin, W.~Yang, B.~Huang, K.~Van~Wyk, H.~Su, X.~Wang, Y.-W. Chao, and D.~Fox, ``Anyteleop: A general vision-based dexterous robot arm-hand teleoperation system,'' in {\em RSS}, 2023.

\bibitem{qin2022one}
Y.~Qin, H.~Su, and X.~Wang, ``From one hand to multiple hands: Imitation learning for dexterous manipulation from single-camera teleoperation,'' {\em RA-L}, vol.~7, no.~4, pp.~10873--10881, 2022.

\bibitem{duan2023ar2}
J.~Duan, Y.~R. Wang, M.~Shridhar, D.~Fox, and R.~Krishna, ``Ar2-d2: Training a robot without a robot,'' in {\em CoRL}, 2023.

\bibitem{qin2022dexmv}
Y.~Qin, Y.-H. Wu, S.~Liu, H.~Jiang, R.~Yang, Y.~Fu, and X.~Wang, ``Dexmv: Imitation learning for dexterous manipulation from human videos,'' in {\em ECCV}, 2022.

\bibitem{fan2022minedojo}
L.~Fan, G.~Wang, Y.~Jiang, A.~Mandlekar, Y.~Yang, H.~Zhu, A.~Tang, D.-A. Huang, Y.~Zhu, and A.~Anandkumar, ``Minedojo: Building open-ended embodied agents with internet-scale knowledge,'' in {\em NeurIPS}, 2022.

\bibitem{kwon2023reward}
M.~Kwon, S.~M. Xie, K.~Bullard, and D.~Sadigh, ``Reward design with language models,'' {\em arXiv preprint arXiv:2303.00001}, 2023.

\bibitem{du2023guiding}
Y.~Du, O.~Watkins, Z.~Wang, C.~Colas, T.~Darrell, P.~Abbeel, A.~Gupta, and J.~Andreas, ``Guiding pretraining in reinforcement learning with large language models,'' in {\em ICML}, 2023.

\bibitem{sermanet2016unsupervised}
P.~Sermanet, K.~Xu, and S.~Levine, ``Unsupervised perceptual rewards for imitation learning,'' in {\em RSS}, 2017.

\bibitem{schmeckpeper2020reinforcement}
K.~Schmeckpeper, O.~Rybkin, K.~Daniilidis, S.~Levine, and C.~Finn, ``Reinforcement learning with videos: Combining offline observations with interaction,'' {\em arXiv preprint arXiv:2011.06507}, 2020.

\bibitem{chang2024look}
M.~Chang, A.~Prakash, and S.~Gupta, ``Look ma, no hands! agent-environment factorization of egocentric videos,'' in {\em NeurIPS}, 2024.

\bibitem{wu2022vat}
R.~Wu, Y.~Zhao, K.~Mo, Z.~Guo, Y.~Wang, T.~Wu, Q.~Fan, X.~Chen, L.~Guibas, and H.~Dong, ``Vat-mart: Learning visual action trajectory proposals for manipulating 3d articulated objects,'' in {\em ICLR}, 2022.

\bibitem{xu2022universal}
Z.~Xu, Z.~He, and S.~Song, ``Universal manipulation policy network for articulated objects,'' {\em RA-L}, vol.~7, no.~2, pp.~2447--2454, 2022.

\bibitem{jiang2022ditto}
Z.~Jiang, C.-C. Hsu, and Y.~Zhu, ``Ditto: Building digital twins of articulated objects from interaction,'' in {\em CVPR}, 2022.

\bibitem{bharadhwaj2023zero}
H.~Bharadhwaj, A.~Gupta, S.~Tulsiani, and V.~Kumar, ``Zero-shot robot manipulation from passive human videos,'' {\em arXiv preprint arXiv:2302.02011}, 2023.

\bibitem{zhang2023flowbot++}
H.~Zhang, B.~Eisner, and D.~Held, ``Flowbot++: Learning generalized articulated objects manipulation via articulation projection,'' {\em arXiv preprint arXiv:2306.12893}, 2023.

\bibitem{shao2020unigrasp}
L.~Shao, F.~Ferreira, M.~Jorda, V.~Nambiar, J.~Luo, E.~Solowjow, J.~A. Ojea, O.~Khatib, and J.~Bohg, ``Unigrasp: Learning a unified model to grasp with multifingered robotic hands,'' {\em RA-L}, vol.~5, no.~2, pp.~2286--2293, 2020.

\bibitem{liu2021synthesizing}
T.~Liu, Z.~Liu, Z.~Jiao, Y.~Zhu, and S.-C. Zhu, ``Synthesizing diverse and physically stable grasps with arbitrary hand structures using differentiable force closure estimator,'' {\em RA-L}, vol.~7, no.~1, pp.~470--477, 2021.

\bibitem{brahmbhatt2019contactgrasp}
S.~Brahmbhatt, A.~Handa, J.~Hays, and D.~Fox, ``Contactgrasp: Functional multi-finger grasp synthesis from contact,'' in {\em IROS}, 2019.

\bibitem{xu2023open}
J.~Xu, S.~Liu, A.~Vahdat, W.~Byeon, X.~Wang, and S.~De~Mello, ``Open-vocabulary panoptic segmentation with text-to-image diffusion models,'' in {\em CVPR}, 2023.

\bibitem{li2022towards}
Z.~Li, C.-Z. Lu, J.~Qin, C.-L. Guo, and M.-M. Cheng, ``Towards an end-to-end framework for flow-guided video inpainting,'' in {\em CVPR}, 2022.

\bibitem{sermanet2018time}
P.~Sermanet, C.~Lynch, Y.~Chebotar, J.~Hsu, E.~Jang, S.~Schaal, S.~Levine, and G.~Brain, ``Time-contrastive networks: Self-supervised learning from video,'' in {\em ICRA}, 2018.

\bibitem{tan2011stable}
J.~Tan, K.~Liu, and G.~Turk, ``Stable proportional-derivative controllers,'' {\em IEEE Computer Graphics and Applications}, vol.~31, no.~4, pp.~34--44, 2011.

\bibitem{fang2020graspnet}
H.-S. Fang, C.~Wang, M.~Gou, and C.~Lu, ``Graspnet-1billion: A large-scale benchmark for general object grasping,'' in {\em CVPR}, 2020.

\bibitem{schulman2017proximal}
J.~Schulman, F.~Wolski, P.~Dhariwal, A.~Radford, and O.~Klimov, ``Proximal policy optimization algorithms,'' {\em arXiv preprint arXiv:1707.06347}, 2017.

\bibitem{he2016deep}
K.~He, X.~Zhang, S.~Ren, and J.~Sun, ``Deep residual learning for image recognition,'' in {\em CVPR}, 2016.

\bibitem{spearman1987proof}
C.~Spearman, ``The proof and measurement of association between two things,'' {\em The American Journal of Psychology}, vol.~100, no.~3/4, pp.~441--471, 1987.

\bibitem{radford2021learning}
A.~Radford, J.~W. Kim, C.~Hallacy, A.~Ramesh, G.~Goh, S.~Agarwal, G.~Sastry, A.~Askell, P.~Mishkin, J.~Clark, {\em et~al.}, ``Learning transferable visual models from natural language supervision,'' in {\em ICML}, 2021.

\bibitem{shah2021rrl}
R.~Shah and V.~Kumar, ``Rrl: Resnet as representation for reinforcement learning,'' in {\em ICML}, 2021.

\bibitem{peng2019advantage}
X.~B. Peng, A.~Kumar, G.~Zhang, and S.~Levine, ``Advantage-weighted regression: Simple and scalable off-policy reinforcement learning,'' {\em arXiv preprint arXiv:1910.00177}, 2019.

\end{thebibliography}
}

\end{document}